\documentclass[runningheads]{llncs}
\usepackage{graphicx}
\usepackage{amsmath,amssymb} % define this before the line numbering.
\usepackage{color}
\usepackage[width=122mm,left=12mm,paperwidth=146mm,height=193mm,top=12mm,paperheight=217mm]{geometry}
\usepackage[percent]{overpic}

\usepackage{listings}
\usepackage{mdframed} 
\usepackage{array,multirow}
\newcommand{\ve}[1]{\mathbf{#1}} % for displaying a vector
\begin{document}
\pagestyle{headings}
\mainmatter

\title{Modality Distillation with Multiple Stream Networks for Action Recognition} % 

%\titlerunning{ECCV-18 submission ID \ECCV18SubNumber}
%\authorrunning{ECCV-18 submission ID \ECCV18SubNumber}
%\author{Anonymous ECCV submission}
%\institute{Paper ID \ECCV18SubNumber}

\author{Nuno C. Garcia$^1$$^2$, Pietro Morerio$^1$, Vittorio Murino$^1$$^3$}
\institute{$^1$Istituto Italiano di Tecnologia $^2$Universita` degli Studi di Genova \\ $^3$Universita` di Verona}

\maketitle

\begin{abstract}
Diverse input data modalities can provide complementary cues for several tasks, usually leading to more robust algorithms and better performance.
However, while a (training) dataset could be accurately designed to include a variety of sensory inputs, 
it is often the case that not all modalities could be  available in real life (testing) scenarios, where a model has to be deployed.
This raises the challenge of how to learn robust representations leveraging multimodal data in the training stage, while considering limitations at test time, such as noisy or missing modalities.

This paper presents a new approach for multimodal video action recognition, developed within the unified frameworks of distillation and privileged information, named generalized distillation.
Particularly, we consider the case of learning representations from depth and RGB videos, while relying on RGB data only at test time. We propose a new approach to train an hallucination network that learns to distill depth features through multiplicative connections of spatiotemporal representations, leveraging soft labels and hard labels, as well as distance between feature maps.
We report state-of-the-art results on video action classification on the largest multimodal dataset available for this task, the NTU RGB+D. Code available at \url{https://github.com/ncgarcia/modality-distillation}
\keywords{action recognition; deep multimodal learning;
distillation; privileged information.}
\end{abstract}

%%%%%%%%%%%%%%%%%%%%%%%%%%%%%%%%%%%%%%%%%%%%%%%%%%%%%%%%%%%%%%%%%
%%%%%%%%%%------------------------------------------%%%%%%%%%%%%%
%%%%%%%%%%%%%%%%%%%%%%%%%%%%%%%%%%%%%%%%%%%%%%%%%%%%%%%%%%%%%%%%%
\section{Introduction}

Imagine to have a large multimodal dataset to train a deep learning model on, for example consisting in RGB video sequences, depth maps, infrared, and skeleton joints data. However, at test time, this model may be used in scenarios where not all of these modalities are available - for example, most of the cameras capture RGB only, which is the most common and cheapest available data modality.

Considering this limitation, what is the best way of using all data available to learn robust representations to be exploited when there are missing modalities at test time? 
In other words, is there any added value to train a model by exploiting more data modalities, even if only one can be used at test time?
The simplest and most commonly adopted solution could be to train the model using only the modality in which it will be tested.
However, a more interesting alternative is trying to exploit the potential of the available data and train the model using all available modalities, realizing, however, that not all of them will be accessible at test time. This learning paradigm, i.e., when the model is trained using extra information, is generally known as \textit{learning with privileged information} \cite{vapnik2009new} or \textit{learning with side information} \cite{hoffman2016learning}. 

In this work, we propose a multimodal stream framework that learns from different data modalities and can be deployed and tested on a subset of these. We design a model able to learn from RGB \textit{and} depth video sequences, but due  to its general structure, it can also be used to manage whatever combination of other modalities as well. To show its potential, we evaluate the performance on the task of video action recognition. In this context, we introduce a new learning paradigm, depicted in Fig. \ref{fig:idea}, to \textit{distill} the information conveyed by depth into an \textit{hallucination} network, which is meant to ``mimic'' the missing stream at test time. Distillation \cite{hinton2014distilling}\cite{ba2014deep} refers to any training procedure where knowledge is transferred from a previously trained complex model to a simpler one. Our learning procedure also introduces a new loss function which is inspired to the \textit{generalized distillation} framework \cite{lopez2015unifying}, that unifies distillation and privileged information learning theories. 
Our model is inspired to the two-stream network introduced by Simonyan and Zisserman \cite{simonyan2014two}, which uses RGB and optical flow, and has been notably successful in the traditional setting for video action recognition task \cite{carreira2017quo}\cite{feichtenhofer2017spatiotemporal}. Differently, we use multimodal data, deploying one stream for each modality (RGB and depth in our case), and use it in the framework of privileged information. 

Another inspiring work is \cite{hoffman2016learning}, which proposed an hallucination network to learn with side information.
We build on this idea, notably extending it by devising a new mechanism to \emph{learn} and \emph{use} such hallucination stream through a more general loss function and inter-stream connections. 

To summarize, the main contributions of this paper are:
\begin{itemize}
\item we propose a new multimodal stream network architecture able to exploit multiple data modalities at training while using only one at test time;
\item we introduce a new learning paradigm to learn an hallucination network within a novel two-stream model;
\item in this context, we have designed an inter-stream connection mechanism to improve the learning process of the hallucination network, and a general loss function, based on the generalized distillation framework;
\item we report state-of-the-art results -- in the privileged information scenario -- on the largest multimodal dataset for video action recognition, the NTU RGB+D \cite{shahroudy2016ntu}. 
\end{itemize}

The rest of the paper is organized as follows. Section \ref{sec:rela} reviews similar approaches and discusses how they relate to the present work. Section \ref{sec:model} details the proposed architecture and the novel learning paradigm. Section \ref{sec:exp} reports the results obtained on the NTU dataset, including a detailed ablation study and a comparative performance with respect to the state of the art. 
Finally, we  draw conclusions and future research directions in section \ref{sec:concl}.

\section{Related Work} \label{sec:rela}

Our work is at the intersection of three topics: privileged information \cite{vapnik2009new}, network distillation \cite{hinton2014distilling}, and multimodal video action recognition. However, Lopez \emph{et al.} \cite{lopez2015unifying} noted that privileged information and network distillation are instances of a the same more inclusive theory, called generalized distillation.

\textbf{Generalized Distillation.}
In \cite{lopez2015unifying}, the unification of two distinct theories related to the concept of machines-teaching-machines is proposed: \emph{privileged information} \cite{vapnik2009new} and \emph{network distillation} \cite{hinton2014distilling}\cite{ba2014deep}. The former, mainly known as learning with privileged information paradigm, proposes to introduce into the learning process a ``teacher'' that supplies additional information to a ``student'' (model or network). The intuition is that the teacher's additional explanations enable the student to learn a better model. Importantly, the additional information provided by the teacher is only available to the student at training time, thus the term \emph{privileged} information.

On the other hand, distillation proposes a training procedure to transfer knowledge from a previously trained large model or ensemble of models to a small model, thus distilling information from a heavier to a lighter model. This idea comes from realizing that the speed and computation requirements for training and testing phases are very different. 

Within the generalized distillation framework, our model is both related to the privileged information theory \cite{vapnik2009new}, considering that the extra modality (depth, in this case) is only used at training time, and, mostly, to the distillation framework. In fact, the core mechanism that our model uses to learn the hallucination network is derived from a distillation loss. 
More specifically, the supervision information provided by the teacher network (in this case, the network processing the depth data stream) is distilled into the hallucination network leveraging teacher's soft predictions and hard ground-truth labels in the loss function.

In this context, the closest works to our proposal are \cite{luo2017graph} and \cite{hoffman2016learning}.
Luo \emph{et. al.} \cite{luo2017graph} addressed a similar problem to ours, where the model is first trained on several modalities (RGB, depth, joints and infrared), but tested only in one. A graph-based distillation method able to distill information from all modalities at training time is proposed, while also passing through a validation phase on a subset of modalities. This showed to reach state-of-the-art results in action recognition and action detection tasks. 
In this respect, our work substantially differs from \cite{luo2017graph} since we benefit from an hallucination mechanism, consisting in an auxiliary network trained using the guidance distilled by the \emph{teacher} network (that processes the depth data stream in our case). Namely, we learn to emulate the presence of the missing modality at test time.

The work of Hoffman \emph{et al.} \cite{hoffman2016learning} introduced a model to hallucinate depth features from RGB input for object detection task. While the idea of using an hallucination stream is similar to the one thereby presented, the mechanism used to learn it is different. \cite{hoffman2016learning} uses an Euclidean loss between depth and hallucinated feature maps, that is part of the total loss along with more than ten classification and localization losses, which makes its effectiveness not very clear since very dependent on hyperparmeter tuning, as the model is trained jointly in one step by optimizing the aforementioned total loss. 

Differently, we propose a loss inspired to the distillation framework, that not only uses the Euclidean distance between feature maps, but also leverages soft predictions from the depth network. Moreover, we encourage the hallucination learning by design, by using cross-stream connections (see Sect. \ref{sec:model}). 
This showed to largely improve the performance of our model with respect to the one-step learning process proposed in \cite{hoffman2016learning}.

\textbf{Multimodal Video Action Recognition.} 
Video action recognition has a long and rich field of literature, spanning from classification methods using handcrafted features to modern deep learning approaches, using either RGB-only or various multimodal data. 
Here, we report some of the more relevant works in multimodal video action recognition, including state-of-the-art methods considering the NTU RGB+D dataset, as well as architectures related to our proposed model.

The two-stream model introduced by Simonyan and Zisserman \cite{simonyan2014two} is a landmark on video analysis, and since then has inspired a series of variants that achieved state-of-the-art performance on diverse datasets. This architecture is composed by an RGB and an optical flow stream, which are trained separately, and then fused at the prediction layer. 
The current state of the art in video action recognition \cite{carreira2017quo} is inspired by such model, featuring 3D convolutions to deal with the temporal dimension, instead of the original 2D ones. 
In \cite{feichtenhofer2017spatiotemporal}, a further  variation of the two-stream approach is proposed, which models spatiotemporal features by injecting the motion stream's signal into the residual unit of the appearance stream. The idea of combining the two streams have also been explored previously by the same authors in \cite{feichtenhofer2016convolutional}. 
\\
Our architecture takes these works as a basis, and shares the idea of connecting the streams via multiplicative connections, although using depth instead of optical flow. Moreover, since we use this mechanism 
in the generalized distillation framework, we realized that employing cross-stream connections helps the distillation process and allows to learn a better hallucination network. The intuition is that the hallucination stream, in learning how to hallucinate features \textit{on average} (i.e. by minimizing the Euclidean distance between feature maps) can strongly benefit from spatial signals which, during training, drive the weight updates at all layers.

In \cite{shahroudy2017deep}, the complementary properties of RGB and depth data are explored, taking the NTU RGB+D dataset as testbed. This work designed a deep autoencoder architecture and a structured sparsity
learning machine, and showed to achieve state-of-the-art results for action recognition. Liu \emph{et al.} \cite{liu2017viewpoint} also use RGB and depth complementary information to devise a method for viewpoint invariant action recognition. Here, dense trajectories from RGB data are first extracted, which are then encoded in viewpoint invariant deep features, while a similar procedure is followed for the depth stream. The RGB and depth features are then used as a dictionary to predict the test label. 

All these previous methods exploited the rich information conveyed by the multimodal data to improve recognition. Our work, instead, proposes a fully convolutional model that exploits RGB and depth data at training time only, and uses exclusively RGB data as input at test time, reaching performance comparable to those utilizing the complete set of modalities in both stages.

%%%%%%%%%%%%%%%%%%%%%%%%%%%%%%%%%%%%%%%%%%%%%%%%%%%%%%%%%%%%%%%%%
%%%%%%%%%%------------------------------------------%%%%%%%%%%%%%
%%%%%%%%%%%%%%%%%%%%%%%%%%%%%%%%%%%%%%%%%%%%%%%%%%%%%%%%%%%%%%%%%
\section{Generalized Distillation with Multiple Stream Networks}\label{sec:model}
\vspace{-0.5em}
\subsection{Model}
\begin{figure}[t!]
	\begin{center}
		\includegraphics[width=0.95\linewidth]{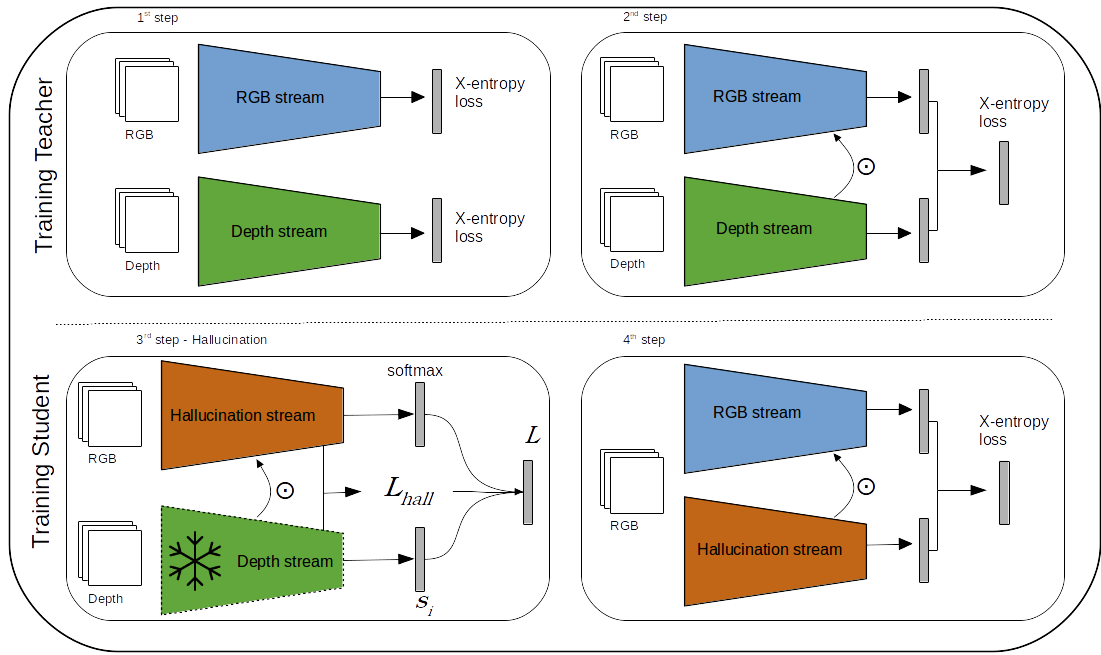}
	\end{center}
	\caption{\footnotesize{Training procedure described in section \ref{sec:training} (see also text therein). The $1^{st}$ step represents the segregate training of the appearance and depth stream networks. The 2nd step illustrates the two-stream joint training.
    The $3^{rd}$ step refers to the hallucination learning step using the soft labels with temperature $s_i$ (eq. \ref{eq:softlabels}) and the novel distillation loss $L$ (eq. \ref{eq:ours}), where the weights of the depth stream network are frozen. The $4^{th}$ step refers to a fine-tuning step, and exemplifies also the testing setup, in which RGB data is the only input to the model.}
    }
	\label{fig:idea}
\end{figure}

\vspace{0em}
\textbf{Cross-stream multiplier networks.}
We design our model (Figure \ref{fig:idea}) based on the architecture presented in \cite{feichtenhofer2017spatiotemporal}, which in turn derives from the two-stream architecture originally proposed in \cite{simonyan2014two}. Typically, the two streams are trained separately and the predictions are fused with a late fusion mechanism. These models use as input appearance (RGB) and motion (optical flow) data, which are fed separately into each stream, both in training and testing. Instead, in this paper we use RGB and depth frames as inputs for training, but only RGB at test time, as already discussed. 

\begin{figure}[t!]
	\begin{center}
		\includegraphics[width=0.8\linewidth]{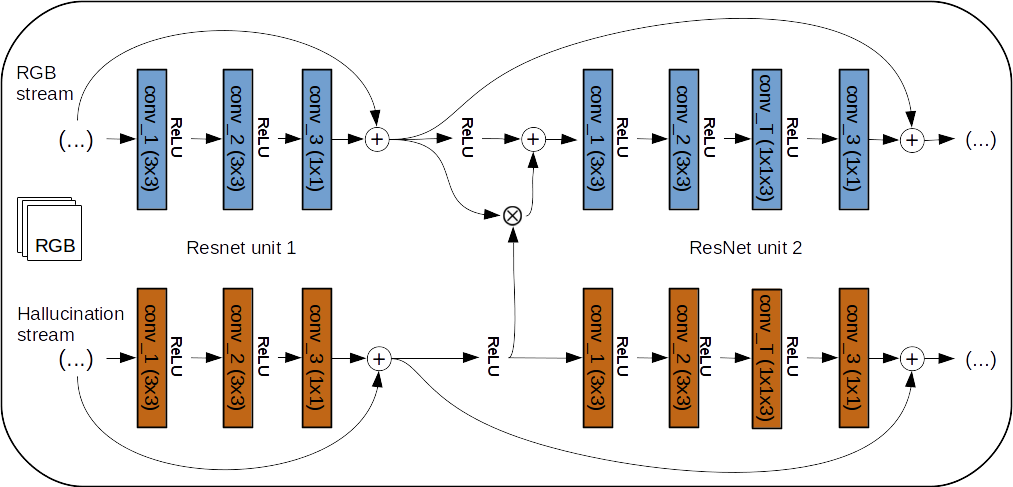}
	\end{center}
	\caption{\footnotesize{Detail of the ResNet residual unit, showing the multiplicative connections and temporal convolutions \cite{feichtenhofer2017spatiotemporal}. In our architecture, the signal injection occurs before the $2^{nd}$ residual unit of each of the four ResNet blocks.}}
	\label{fig:resnetdetail}
\end{figure}

We use the ResNet-50-based \cite{he2016deep}\cite{he2016identity} model proposed in \cite{feichtenhofer2017spatiotemporal} as baseline architecture for each stream block of our model.
In this paper, Feichtenhofer \emph{et al.} proposed to connect the appearance and motion streams with multiplicative connections at several layers, as opposed to previous models which would only interact at the prediction layer. Such connections are depicted in Figure \ref{fig:idea} with the $\odot$ symbol.
Figure \ref{fig:resnetdetail} illustrates this mechanism at a given layer of the multiple stream  architecture, but, in our work, it is actually implemented at the four convolutional layers of the Resnet-50 model. The underlying intuition is that these connections enable the model to learn better spatiotemporal representations, 
and help to distinguish between identical actions that require the combination of appearance and motion features.
Originally, the cross-stream connections consisted in the injection of the motion stream signal into the other stream's residual unit, without affecting the skip path. ResNet's residual units are formally expressed as:
\begin{gather}
\ve{x}_{l+1} = f(h(\ve{x}_{l}) + F(\ve{x}_{l}, \mathcal{W}_l)),
\end{gather}
where $\ve{x}_{l}$ and $\ve{x}_{l+1}$ are $l$-th layer's input and output, respectively, $F$ represents the residual convolutional layers defined by weights $\mathcal{W}_l$, $h(\ve{x}_{l})$ is an identity mapping and $f$ is a ReLU non-linearity.
The cross-streams connections are then defined as
\begin{gather}
\ve{x}^a_{l+1} = f(\ve{x}^a_{l}) + F(\ve{x}^a_{l} \odot f(\ve{x}^m_{l}), \mathcal{W}_l),
\end{gather}
where $\ve{x}^a$ and $\ve{x}^m$ are the appearance and motion streams, respectively, and $\odot$ is the element-wise multiplication operation.
Such mechanism implies a spatial alignment between both feature maps, and therefore between both modalities. This alignment comes for free when using RGB and optical flow, since the latter is computed from the former in a way that spatial arrangement is preserved. However, this is an assumption we can not generally made. For instance, depth and RGB are often captured from different sensors, likely resulting in spatially misaligned frames. We cope with this alignment problem in the method's initialization phase (described in the supplementary material).
In order to augment the model temporal support, 1D temporal convolutions into the second residual unit of each ResNet layer is also included \cite{feichtenhofer2017spatiotemporal}, as illustrated in Fig. \ref{fig:resnetdetail}. The weights $W_{l} \in \mathbb{R}^{1 \times 1 \times 3 \times C_{l} \times C_{l}}$ are convolutional filters initialized as identity mappings at feature level, and centered in time, and $C_{l}$ are the number of channels in layer $l$.

\subsubsection{Hallucination stream.}
We also introduce and learn a hallucination network \cite{hoffman2016learning}, using a new learning paradigm, loss function and design mechanism.
The hallucination stream network has the same architecture as the appearance and depth stream models. 
This network receives RGB as input, and is trained to ``imitate'' the depth stream at different levels, \emph{i.e.} at feature and prediction layers. In this paper, we explore several ways to implement such learning paradigm, including both the training procedure and the loss, and how they affect the overall performance of the model.

In \cite{hoffman2016learning}, a regression loss between the hallucination and depth feature maps is designed, defined as:
\begin{gather}
\vspace{-10pt}
\label{eq:hall}
L_{hall}(l) = \lambda_l \|\sigma(A_l^d)- \sigma(A_l^h)\|^2_2 ,
\vspace{-10pt}
\end{gather}
where $\sigma$ is the sigmoid function, and $A_l^d$ and $A_l^h$ are the l-th layer activations of depth and hallucination network. This Euclidean loss forces both activation maps to be similar. In \cite{hoffman2016learning}, this loss is weighted along with another ten classification and localization loss terms, making it hard to balance the total loss. One of the main motivations behind our proposed new staged learning paradigm, described in section \ref{sec:training},
is to avoid the inefficient, heuristic-based tweaking of so many loss weights, aka hyper-parameter tuning.

Instead, we adopt an approach inspired by the generalized distillation framework \cite{lopez2015unifying}, in which a \textit{student} model $f_s \in \mathcal{F}_s$ distills the representation $f_t \in \mathcal{F}_t$ learned by the \textit{teacher} model.
This is formalized as
\begin{gather}
\vspace{-10pt}
\label{eq:distill}
f_s = \operatorname*{arg\,min}_{f\in\mathcal{F}_s}
\frac{1}{n}\displaystyle\sum_{i=1}^{n}
L_{GD}(i),    n = 1,...,N
\vspace{-10pt}
\end{gather}
where $N$ is the number of examples in the dataset. The generalized distillation loss is so defined as:
\begin{gather}
\vspace{-10pt} 
\label{eq:loss_GD}
L_{GD}(i) = (1-\lambda) \ell (y_i,\sigma(f(x_i))) + \lambda \ell (s_i, \sigma (f(x_i))), \; \lambda \in [0,1]
\vspace{-10pt} 
\end{gather}
and $s_i$ are the soft predictions from the teacher network, that is:
\begin{gather}
\vspace{-10pt} 
\label{eq:softlabels}
s_i = \sigma(f_t(x_i) / T), \; T>0.
\vspace{-10pt} 
\end{gather}
The parameter $\lambda$ in equation \ref{eq:loss_GD} allows to tune the loss by giving more importance either to imitating hard or soft labels, $y_i$ and $s_i$, respectively, actually improving the transfer of information from the depth (teacher) to the hallucination (student) network. The temperature parameter $T$ in equation \ref{eq:softlabels} allows to smooth the probability vector predicted by the teacher network. The intuition is that such smoothing may expose relations between classes that would not be easily revealed in raw predictions, further facilitating the distillation by the student network $F_s$. 

We suggest that both Euclidean and generalized distillation losses are indeed useful in the learning process. In fact, by encouraging the network to decrease the distance between hallucinated and true depth feature maps, it can help to distill depth information encoded in the generalized distillation loss. Thus, we formalize our final loss function as follows:
\begin{gather}
\label{eq:ours}
L = (1-\alpha) L_{GD} + \alpha L_{hall}, \; \alpha \in [0,1],
\end{gather}

where $\alpha$ is a parameter balancing the contributions of the 2 loss terms during training.
The parameters $\lambda$, $\alpha$ and $T$ are estimated by utilizing a validation set. The details for their setting will be provided in the supplementary material.

In summary, the generalized distillation framework proposes to use the student-teacher framework introduced in the distillation theory to extract knowledge from the privileged information source. We explore this idea by proposing a new learning paradigm to train an hallucination network using privileged information, which we will describe in the next section.
In addition to the loss functions introduced above, we also allow the teacher network to share information with the student network by design, through the cross-stream multiplicative connections. We test how all these possibilities affect the model's performance in the experimental section through an extensive ablation study. 

\subsection{Training Paradigm}
\label{sec:training}
In general, the proposed training paradigm, illustrated in Fig. \ref{fig:idea}, is divided in two core parts: the first part (Step 1 and 2 in the figure) focuses on learning the teacher network $F_t$, leveraging RGB and depth data (the privileged information in this case); the second part (Step 3 and 4 in the figure) focuses on learning the hallucination network, referred to as student network $F_s$ in the distillation framework, using the general hallucination loss defined in Eq. \ref{eq:ours}. 

The \textit{first} training step consists in training both streams separately, which is a common practice in two-stream architectures. Both depth and appearance streams are trained minimizing cross-entropy, after being initialized with a pre-trained ImageNet model for all experiments. As in \cite{eitel2015multimodal}, depth frames are encoded into color images using a jet colormap.

The \textit{second} training step is still focused on further training the teacher model. This step gives the basis for the following hallucination network training, which, receiving in input RGB data, should behaves like an actual depth stream network. 
For this reason, we must train the depth stream network in the same setting as the hallucination model will act, hence, it is trained considering the cross-stream connections and adding the prediction fusion layer with the RGB stream model.
Since the model trained in this step has the architecture and capacity of the final one, and \textit{has access to both modalities}, its performance represents an upper bound for the task we are addressing. 
This is one of the major differences between our approach and the one used in \cite{hoffman2016learning}: by decoupling the teacher learning phase with the hallucination learning, we are able to both learn a better teacher \textit{and} a better student, as we will show in the experimental section.

In the \textit{third} training step, we focus on learning the hallucination network from the teacher model, i.e., the depth stream network just trained. Here, the weights of the depth network are frozen, while receiving in input depth data. Instead, the hallucination network, receiving in input RGB data, is trained 
with the loss defined in \ref{eq:ours}, while also receiving feedback from the cross-stream connections from the depth network. We found that this helps the learning process.

In the \textit{fourth} and last step, we carry out fine tuning of the whole model, composed by the RGB and the hallucination streams. This step uses RGB only as input, and it also precisely resembles the setup used at test time.
The cross-stream connections inject the hallucinated signal into the appearance RGB stream network, resulting in the multiplication of the hallucinated feature maps and the RGB feature maps. The intuition is that the hallucination network has learned to inform the RGB model where the action is taking place, similarly to what the depth model would do with real depth data.

A summary of the whole training process is reported as in the following box:
% \fbox{
% \parbox{\textwidth}{
\begin{scriptsize}
\begin{mdframed}
\begin{itemize}
\item \textbf{training step 1}
\begin{itemize}
\item initialize RGB and depth streams with ImageNet-pretrained weights;
\item train depth and RGB streams \textit{separately}, with depth and RGB data respectively and standard cross entropy classification loss;
\end{itemize}
\item \textbf{training step 2} (\textit{learning the teacher network})
\begin{itemize}
\item initialize both streams with weights learned in step 1;
\item train both streams jointly as a two-stream model \cite{feichtenhofer2017spatiotemporal} (i.e. with multiplier connections), using both RGB and depth data, with cross entropy loss;
\end{itemize}
\item \textbf{training step 3} (\textit{learning the student network})
\begin{itemize}
\item freeze depth network weights learned in step 2;
\item initialize hallucination network with depth weights;
\item train with cross-stream connections and the proposed loss $L$ (eq. \ref{eq:ours});
\end{itemize}
\item \textbf{training step 4} (\textit{finetune the final model})
\begin{itemize}
\item initialize the hallucination stream with weights learned in step 3;
\item initialize RGB stream with weights from step 2;
\item fine-tune the joint model composed by hallucination + RGB branches (with cross-stream connections) using RGB data only and cross entropy loss;
\end{itemize}
\end{itemize}
% }}
\end{mdframed}
\end{scriptsize}

%%%%%%%%%%%%%%%%%%%%%%%%%%%%%%%%%%%%%%%%%%%%%%%%%%%%%%%%%%%%%%%%%
%%%%%%%%%%------------------------------------------%%%%%%%%%%%%%
%%%%%%%%%%%%%%%%%%%%%%%%%%%%%%%%%%%%%%%%%%%%%%%%%%%%%%%%%%%%%%%%%
\begin{figure}
	\centering
	\setlength{\tabcolsep}{2pt}
	\begin{tabular}{ccccc}		
        \includegraphics[width=.15\linewidth]{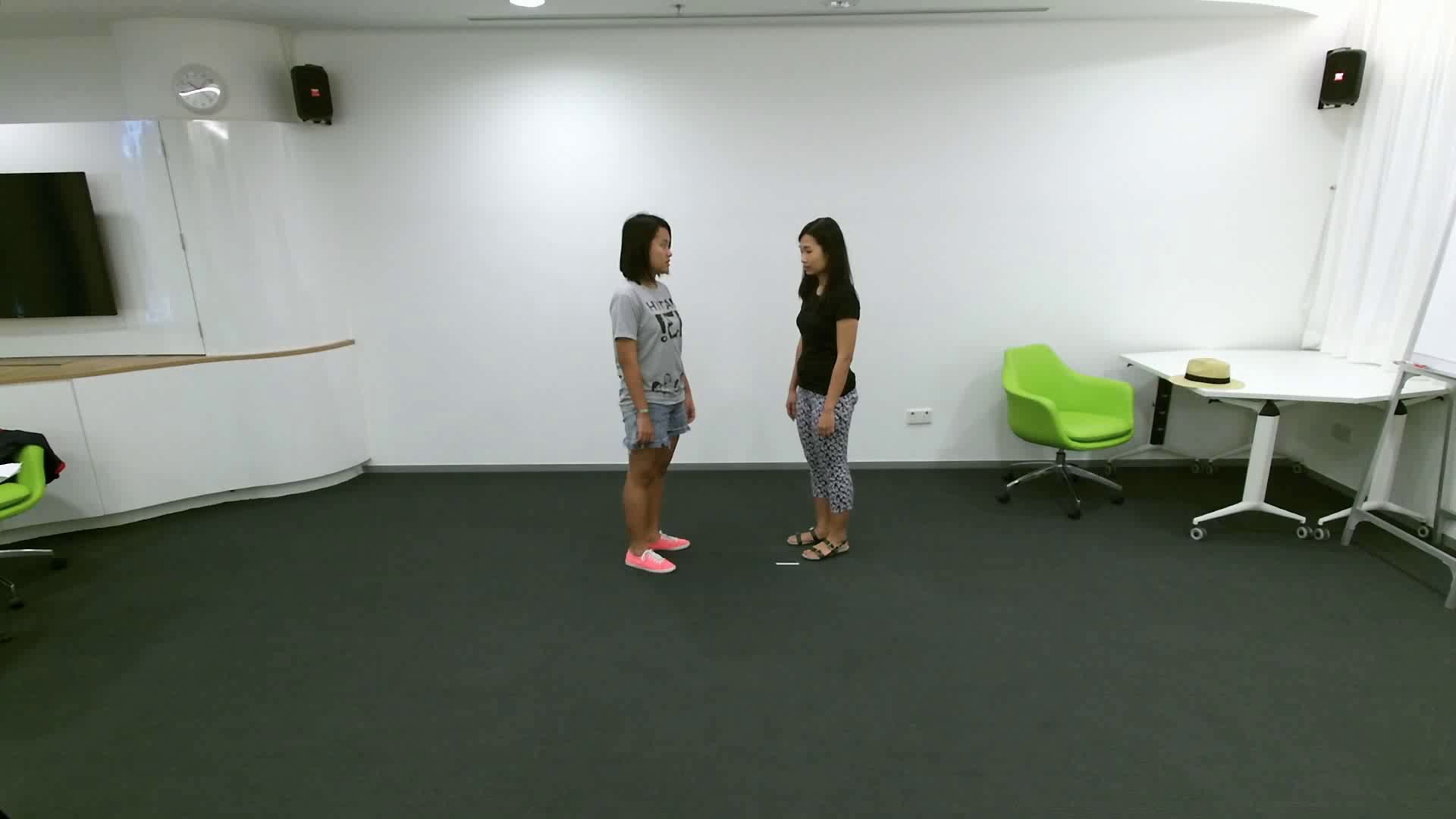} & 
		\includegraphics[width=.15\linewidth]{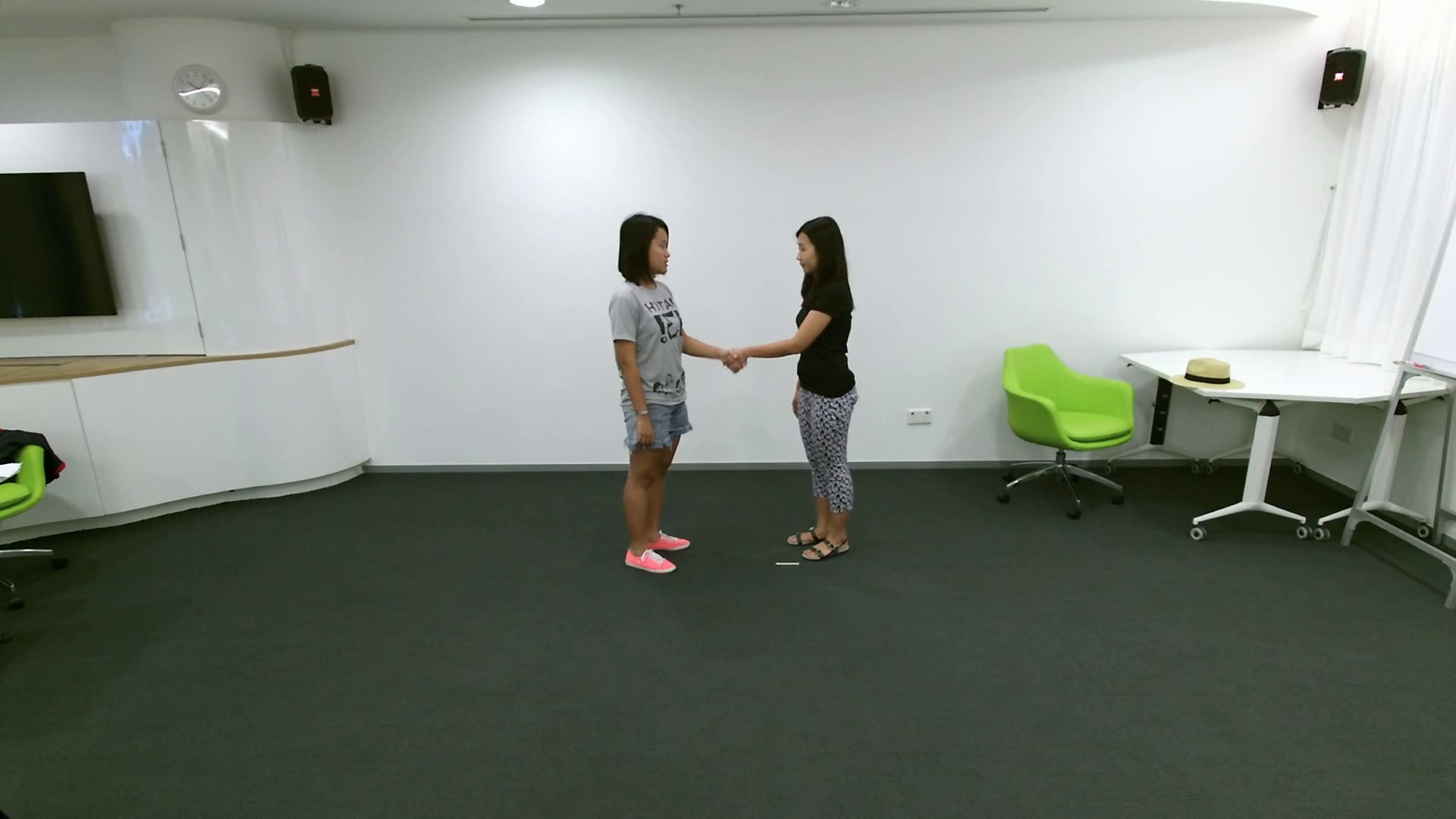} & 
		\includegraphics[width=.15\linewidth]{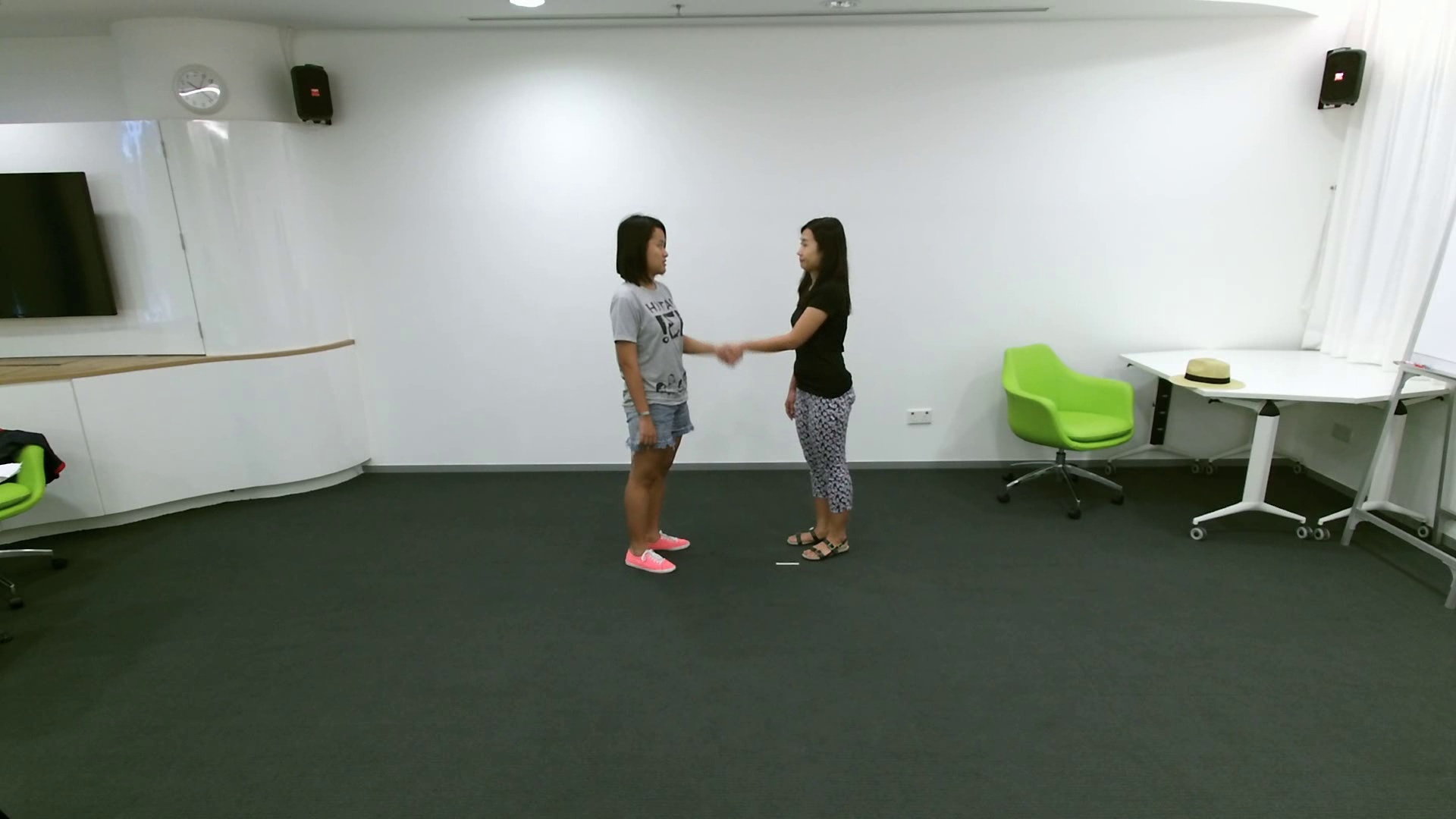} & 
		\includegraphics[width=.15\linewidth]{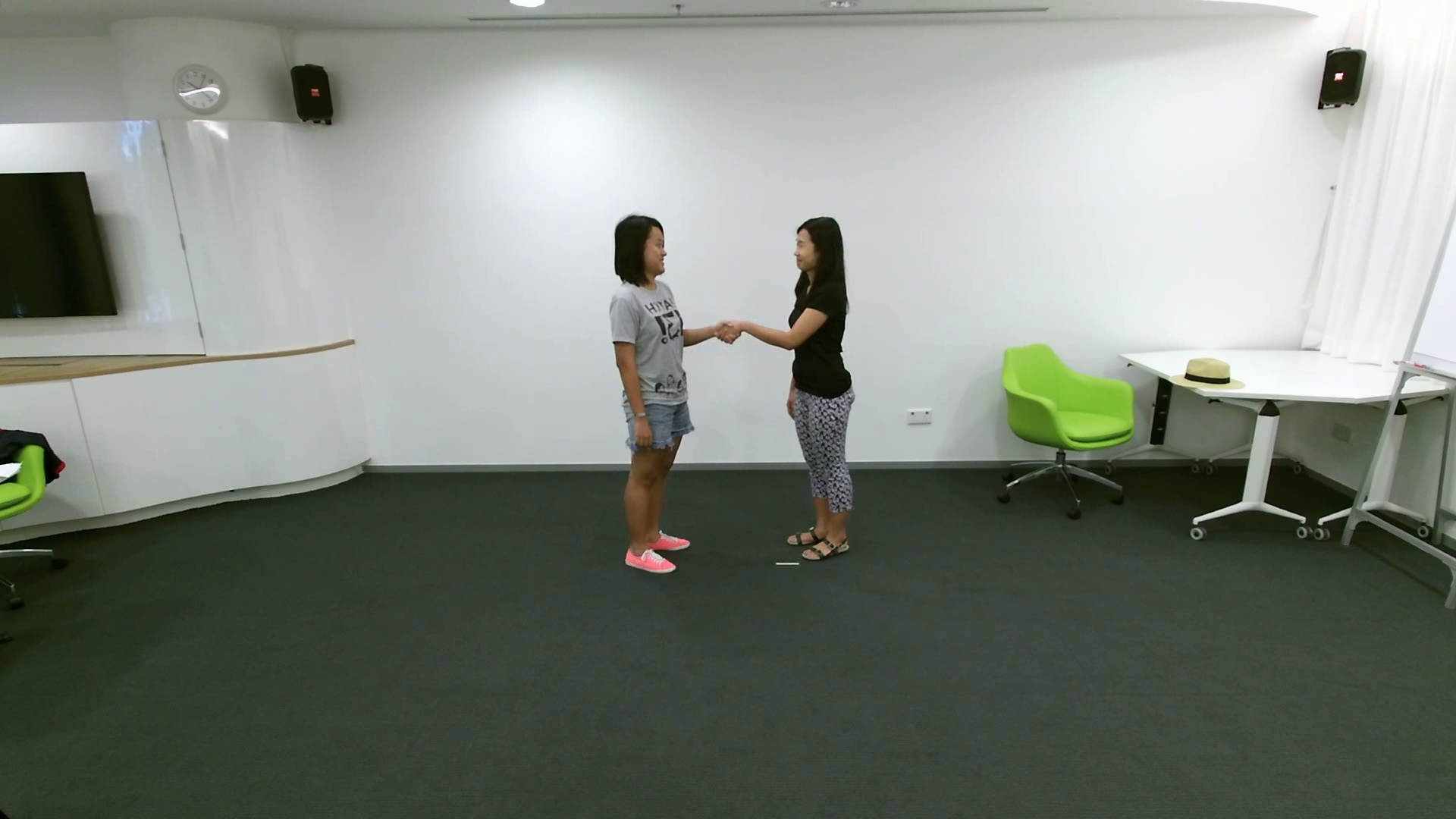} & 
		\includegraphics[width=.15\linewidth]{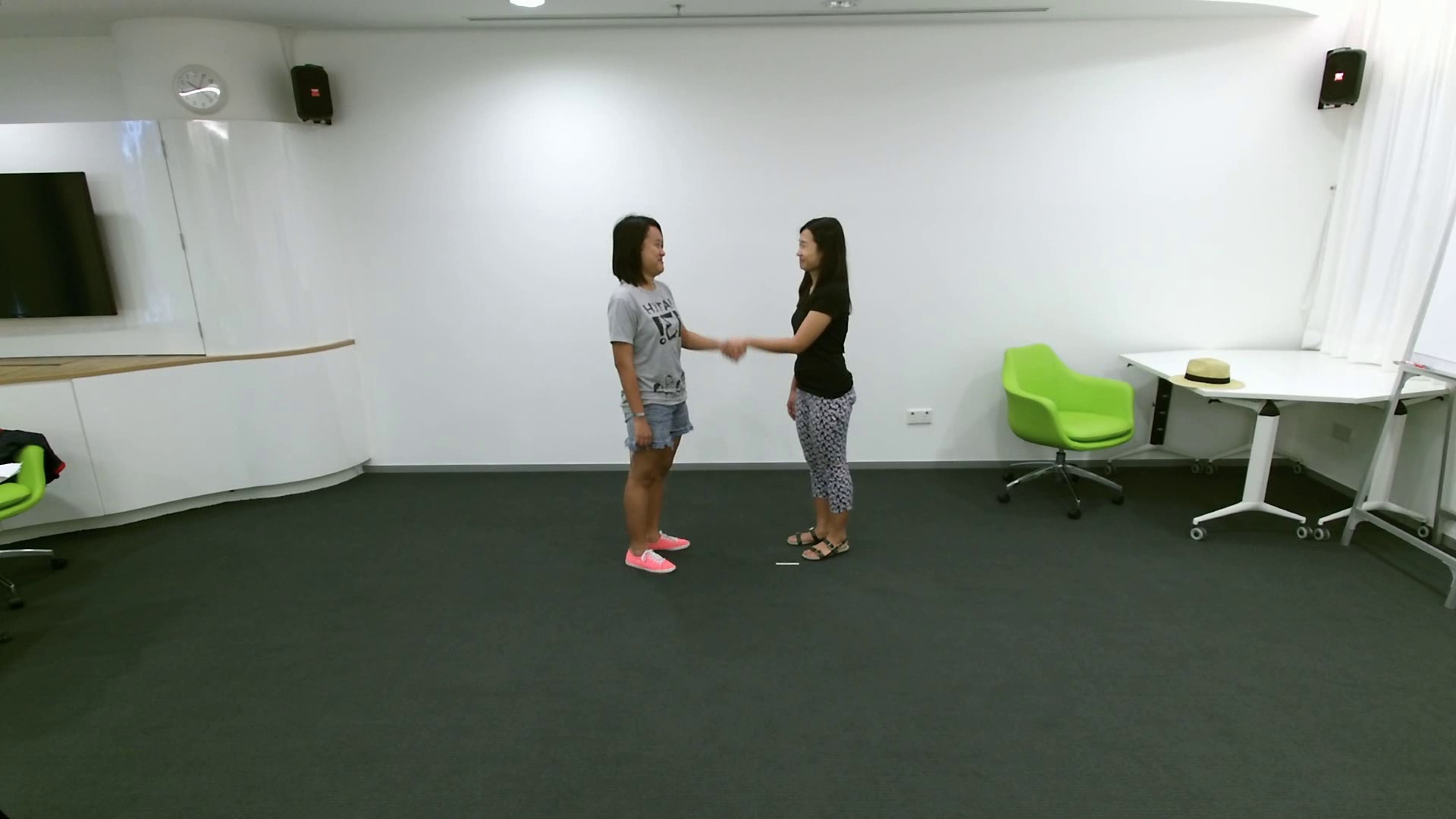} \\
		\includegraphics[width=.15\linewidth]{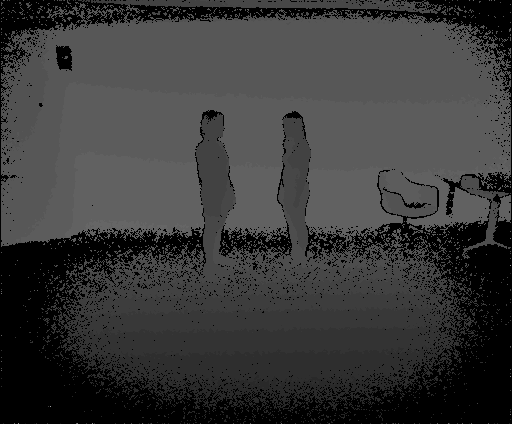} &
        \includegraphics[width=.15\linewidth]{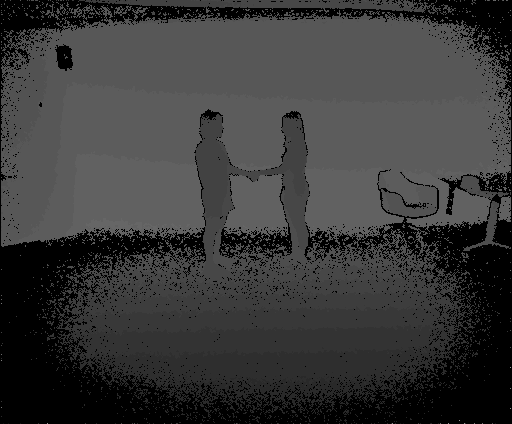} & 
		\includegraphics[width=.15\linewidth]{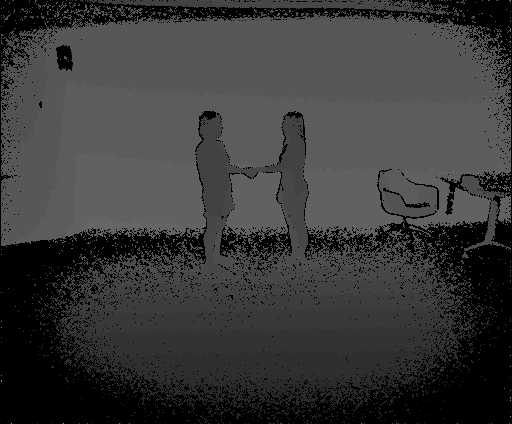} & 
		\includegraphics[width=.15\linewidth]{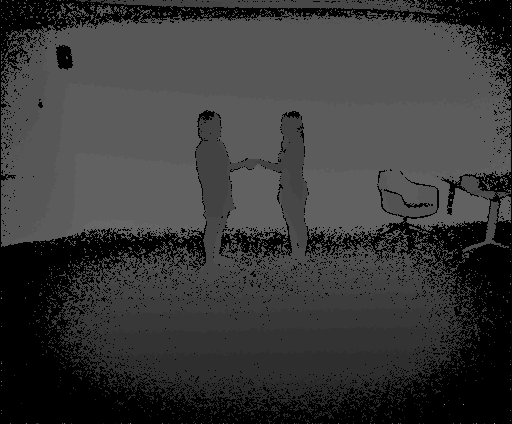} &  
		\includegraphics[width=.15\linewidth]{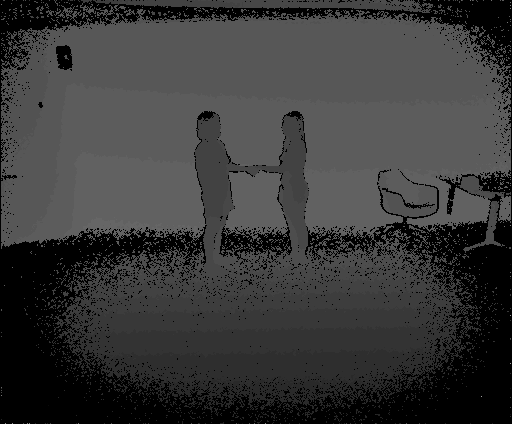}
	\end{tabular}
	\caption{Example of RGB and depth frames from the NTU RGB+D Dataset. }
	\label{fig:ntuframes}
\end{figure}

\section{Experiments}\label{sec:exp}

\subsection{Dataset}
\label{sec:dataset}
We evaluate our model on the NTU RGB+D dataset \cite{shahroudy2016ntu}, which is currently the largest public dataset for multimodal video action recognition. It is composed by 56,880 videos, available in four modalities: RGB videos, depth sequences, infrared frames, and 3D skeleton data of 25 joints. It was acquired with a Kinect v2 sensor in 80 different viewpoints, and includes 40 subjects performing 60 distinct actions, including daily simple actions (\emph{e.g.}, brushing teeth, drinking, writing), interactions (\emph{e.g.}, kicking other person, hugging other person), and health-related actions (\emph{e.g.}, nausea or vomiting condition, sneeze/cough). 
We follow the two evaluation protocols originally proposed in \cite{shahroudy2016ntu}, which are cross-subject and cross-view. As in the original paper, we use about 5\% of the training data as validation set for both protocols, in order to select the parameters $\lambda$, $\alpha$ and $T$. In this paper, we use only RGB and depth data. The masked depth maps are converted to a three channel map via a jet mapping, as in \cite{eitel2015multimodal}.

\subsection{Comparison with state of the art}
\vspace{-1pt}
Table \ref{table:sota1} compares performances of different methods on the NTU RGB+D dataset. Classification accuracy is the standard performance measure used for this dataset: it is estimated according to the protocols (training and testing splits) reported in the respective works we are comparing with.
The first part of the table (indicated by $\times$ symbol) refers to the unsupervised method proposed in \cite{luo2017unsupervised}, which achieve surprisingly high results even without relying on labels in learning representations. 
The second part refers to supervised methods (indicated by $\bigtriangleup$), divided according to the modalities used for training and testing. Here, we list the performance of the separate RGB and depth streams trained in step 1, as a reference. Of course, we expect our final model to perform better than the one trained on RGB only. We also propose our baseline, consisting in the teacher model trained in step 2. Its accuracy represents an upper bound for the final model, which will not rely on depth data at test time.
The last part of the table (indicated by $\Box$) reports our model's performances at 2 different stages together with the other privileged information method \cite{hoffman2016learning}. 
For both protocols, we can see that our privileged information approach outperforms \cite{hoffman2016learning}, which is the only fair \textit{direct} comparison we can make (same training \& test data). Besides, as expected, our final model performs better than ``Ours - RGB model, step 1'' since it exploits more data at training time, and worse than ``Ours - step 2'', since it exploits less data at test time. Other RGB+D methods perform better (which is comprehensible since they rely on RGB+D in both training and test) but not by a large margin. More details and additional comments on the compared methods are provided in the supplementary material.
\begin{table}
	\centering
	\begin{center}
		\begin{tabular}{|l|l|c|c|c|}
			\hline
			Method & Test Modalities & Cross Subject & Cross View &  \\
            \hline
            \hline
			Luo \cite{luo2017unsupervised} & Depth & 66.2\% & - & \multirow{2}{*}{$\times$}\\ 
			Luo \cite{luo2017unsupervised} & RGB & 56.0\% & - &\\
			\hline
			\hline
			HOG-2 \cite{ohn2013joint} & Depth & 32.4\% & 22.3\% & \multirow{10}{*}{$	\bigtriangleup$}\\
            \textbf{Ours - depth model, step 1} & \textbf{Depth} & \textbf{70.44}\% & \textbf{75.16}\% &\\
			\textbf{Ours - RGB model, step 1} & \textbf{RGB} & \textbf{66.52}\% & \textbf{71.39}\% &\\
			Deep RNN \cite{shahroudy2016ntu} & Joints & 56.3\% & 64.1\% &\\
			Deep LSTM \cite{shahroudy2016ntu} & Joints &   60.7\%  & 67.3\% &\\
			Sharoudy \cite{shahroudy2016ntu} & Joints &   62.93\%  & 70.27\% &\\
			Kim \cite{soo2017interpretable} & Joints & 74.3\% & 83.1\% &\\
			Sharoudy \cite{shahroudy2017deep} & RGB+D &  74.86\%  & - &\\
			Liu \cite{liu2017viewpoint} & RGB+D & 77.5\% & 84.5\% &\\
            \textbf{Ours - step 2} & \textbf{RGB+D} & \textbf{79.73}\% & \textbf{81.43}\% &\\
			\hline
			\hline			
            Hoffman \emph{et al.} \cite{hoffman2016learning} & RGB & 64.64\% & - & \multirow{3}{*}{$\Box$}\\
			\textbf{Ours - step 3} & \textbf{RGB} & \textbf{71.93}\% & \textbf{74.10}\% &\\ 
            \textbf{Ours - step 4} & \textbf{RGB} & \textbf{73.42}\% & \textbf{77.21}\% &\\ 
            \hline
		\end{tabular}
	\end{center}
	\caption{Classification accuracies and comparisons with the state of the art. Performances referred to the several steps of our approach (ours) are highlighted in bold. $\times$ refers to comparisons with unsupervised learning methods.  $\bigtriangleup$ refers to supervised methods: here train and test modalities coincide. $\Box$ refers to privileged information methods: here training exploits RGB+D data, while test relies on RGB data only.}
        \vspace{-25pt}
	\label{table:sota1}
\end{table}

\subsection{Ablation study}
\label{sec:ablation}

\begin{table}
	\centering
	\begin{center}
		\begin{tabular}{|ll|l|c|c|c|}
			\hline
            \# & Method & Test Modality & Loss &  Cross-Subject & Cross-View \\
			\hline
            \hline
			1 &Ours - step 1, depth stream & Depth &x-entr& 70.44\% & 75.16\% \\
			2&Ours - step 1, RGB stream & RGB &x-entr& 66.52\% & 71.39\% \\
            \hline \hline
            3&Hoffman \cite{hoffman2016learning} w/o connections & RGB & eq. (\ref{eq:hall}) & 64.64\% & - \\
            4&Hoffman \cite{hoffman2016learning} w/o connections & RGB & eq. (\ref{eq:loss_GD})& 68.60\% & - \\
            5&Hoffman \cite{hoffman2016learning} w/o connections & RGB & eq. (\ref{eq:ours})& 70.70\% & - \\
            \hline \hline
            6&Ours - step 2, depth stream & Depth &x-entr& 71.09\% & 77.30\% \\
            7&Ours - step 2, RGB stream & RGB &x-entr& 66.68\% & 56.26\% \\
			8&Ours - step 2 & RGB \& Depth &x-entr& \textbf{79.73}\% & 81.43\% \\
            9&Ours - step 2 w/o connections & RGB \& Depth &x-entr& 78.27\% & \textbf{82.11}\% \\

			\hline \hline
            10&Ours - step 3 w/o connections & RGB (\textit{hall}) & eq. (\ref{eq:hall}) & 69.93\% & 70.64\% \\
            11&Ours - step 3 w/ connections & RGB (\textit{hall}) & eq. (\ref{eq:hall}) & 70.47\% & - \\
            12&Ours - step 3 w/ connections & RGB (\textit{hall}) & eq. (\ref{eq:distill})& 71.52\% & - \\
			13&Ours - step 3 w/ connections & RGB (\textit{hall}) & eq. (\ref{eq:ours})& 71.93\% & 74.10\% \\
			14&Ours - step 3 w/o connections & RGB (\textit{hall}) & eq. (\ref{eq:ours})& 71.10\% & - \\
            \hline \hline
            15&\textbf{Ours - step 4} & \textbf{RGB}  &x-entr& \textbf{73.42}\% & \textbf{77.21}\% \\
			\hline
		\end{tabular}
	\end{center}
	\caption{Ablation study. A full set of experiments is provided for cross-subject evaluation protocol, and for the cross-view protocol, only the most important results are reported.
    }
    \vspace{-25pt}
	\label{table:ablation}
\end{table}

In this subsection, we discuss the results of the experiments carried out to understand the contribution of each part of the model and of the training procedure.
Table \ref{table:ablation} reports performances at the several training steps, different losses and model configurations. 

Rows \#1 and \#2 refers to the first training step, where depth and RGB streams are trained separately. We can note that the depth stream network provides better performance with respect to the RGB one, as expected. 

The second part of the table (Rows \#3-5) shows the results using Hoffman \emph{et al.}'s method \cite{hoffman2016learning}, \emph{i.e.}, adopting a model initialized with the pre-trained networks from the first training step, and the hallucination network initialized using the depth network. Row \#3 refers to the original paper \cite{hoffman2016learning} (i.e., using the loss $L_{hall}$, Eq. \ref{eq:hall}), and rows \#4 and \#5 refer to the training using the proposed losses $L_{GD}$ and $L$, in Eqs. \ref{eq:loss_GD} and \ref{eq:ours}, respectively. It can be noticed that the accuracies achieved using our proposed loss functions overcome that obtained in \cite{hoffman2016learning} by a significant margin (about 6\% in the case of the total loss $L$).

The third part of the table reports performances after the training step 2. Rows \#6 and \#7 refer to the depth and RGB stream networks belonging to the model of row \#8. This model corresponds to the architecture described in \cite{feichtenhofer2017spatiotemporal} and constitutes the upper bound for our hallucination model, since it uses RGB and depth for training and testing. Performances obtained by the model in row \#8 and \#9, with and without cross-stream connections, respectively, are the highest in absolute when using both modalities (around 78-79\% for cross-subject and 81-82\% for cross-view protocols, respectively), largely outperforming the accuracies obtained using only one modality (in rows \#6 and \#7). 

The fourth part of the table (rows \#10-14) shows results for our hallucination network after the several variations of learning processes, different losses and using or not using the cross-stream connections. One can note that the achieved performances when \textit{only} RGB data are given in input, are in line with those achieved by the model fed by depth data. Depending on the variant adopted, accuracies are around 70-72\%, reaching about 72\% in the case of application of our full model before the fine-tuning step (row \#14, cross-subject protocol). The depth stream model (in row \#6) reaches 71\%, whereas the model with both modalities in input (fixing the upper bound, row \#8) reaches about 79\%: only 6 percentage points separate the 2 models, showing the goodness of our proposed approach.

Finally, the last row, \#15, reports results after the last fine-tuning step, which allows to reach the best accuracy with only the RGB modality as input, increasing the previous performance of about 1.5\%, so narrowing the gap to the upper bound to about 4.5\%.

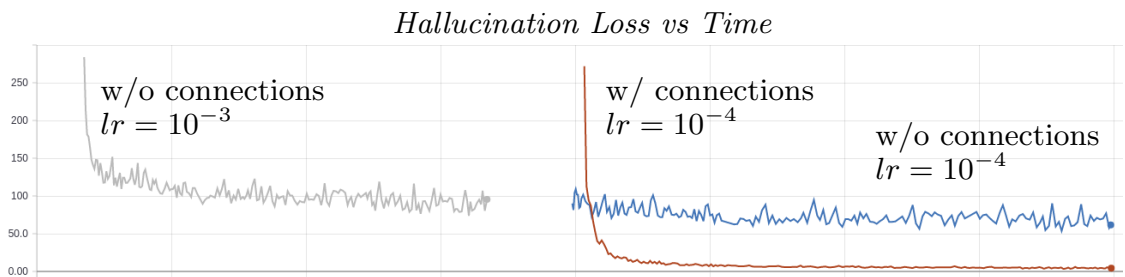
\begin{figure}[b!]
	\begin{center}
		\begin{overpic}[width=0.95\textwidth]{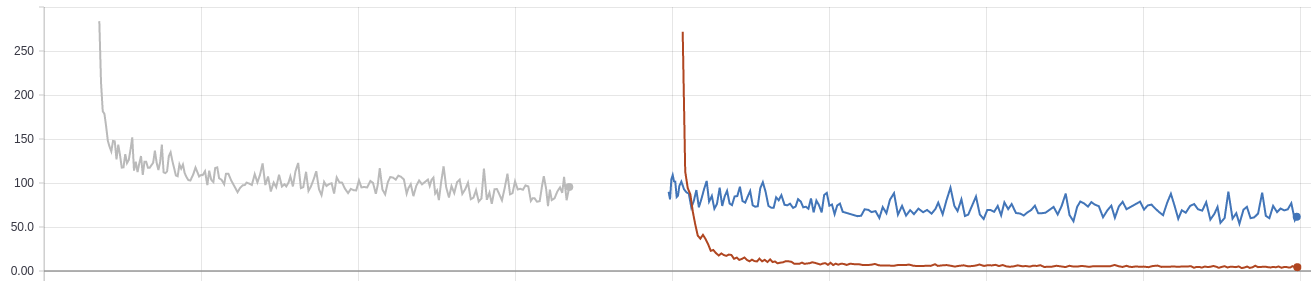}
        \put(9, 16){\footnotesize{w/o connections}}
        \put(9, 13){\footnotesize{$lr=10^{-3}$}}
        
        \put(78, 12){\footnotesize{w/o connections}}
        \put(78, 9){\footnotesize{$lr=10^{-4}$}}
        
        \put(54, 16){\footnotesize{w/ connections}}
        \put(54, 13){\footnotesize{$lr=10^{-4}$}}
        
        \put(35,22){\textit{Hallucination Loss vs Time}}
        \end{overpic}
	\end{center}
    \vspace{-10pt}
	\caption{The plot shows the hallucination loss $L_{hall}$ of Eq. \ref{eq:hall}: the  gray and blue curves refers to the model where no multiplicative connections are used to learn the hallucination stream (row \#14 of Table \ref{table:ablation}). We started the experiment with learning rate set to 0.001, and continued after a while with learning rate set to 0.0001. The red curve shows instead $L_{hall}$ after plugging the inter-stream connections (row \#13  of Table \ref{table:ablation}).}
	\label{fig:loss}
\end{figure}

\textbf{Contribution of the cross-stream connections.} We claim that the signal injection provided by the cross-stream connections helps the learning of a better hallucination network. Row \#13 and \#14 show the performances for the hallucination network learning process, starting from the same point and using the same loss. The hallucination network that is learned using multiplicative connections performs better than its counterpart. This is illustrated in figure \ref{fig:loss}: even after approximately half the number of iterations, the hallucination network learned with the multiplicative cross-stream connections is able to better minimize the Euclidean loss of eq. \ref{eq:hall}. 

\textbf{Contributions of the proposed distillation loss (Eq. \ref{eq:ours}).
}The distillation and Euclidean losses have complementary contributions to the learning of the hallucination network. This is observed by looking at the performances reported in rows \#3, \#4 and \#5, and also \#11, \#12 and \#13. Within both the training procedure proposed by Hoffman \textit{et al.} \cite{hoffman2016learning} and  our staged training process, the distillation loss improves over the Euclidean loss, and the combination of both improves over the rest. This suggests that both Euclidean and distillation losses have its own share and act differently to align the hallucination (student) and depth (teacher) feature maps and outputs' distributions.

\textbf{Contributions of the proposed training procedure.}The intuition behind the staged training procedure proposed in this work can be ascribed to the \textit{dividi et impera} (divide-and-conquer) strategy. In our case, it means breaking the problem in two parts: learning the actual task we aim to solve and learning the hallucination network to face test-time limitations.
Row \#5 reports accuracy for the architecture proposed by Hoffman \textit{et al.}, and rows \#15 report the performance for our model with connections. Both use the same loss to learn the hallucination network, and both start from the same initialization. We observe that our method outperform the one in row \#5, which justifies the proposed staged training procedure.
\\ \\ 
Finally, we motivate for the use of the hallucination model in comparison with other naive approaches when dealing with missing or noisy modalities. Comparing rows \#2 with \#15, we further confirm (if still needed) that using the hallucination model is in fact more useful than training only with RGB data. We also observe that it is more useful to use our hallucination model than naively use totally corrupted depth data as input to the two-stream model. This is observed by comparing results in Table \ref{tab:noise} and the performance at row \#15 in Table \ref{table:ablation}. The following section studies with further detail the behavior of our model when tested using noisy depth data as input.

\subsection{Inference with noisy depth}\label{subsec:noise}
\vspace{-0.5em}

Suppose that in a real test case we can only access unreliable, i.e., noisy, depth data. Now the question is: how much we can trust such data? How better would it be to use a model in which depth is provided by an hallucination network, like that proposed in this work?
In other words, we are finally interested in exploring how our model works under stress, and, more precisely, at which level of noise, hallucinating the depth modality becomes favorable with respect to using the full model with both input modalities (step 2). 
\begin{table}[ht]
	\centering
    \setlength{\tabcolsep}{0em}
		\begin{tabular}{|c||c|ccccc|c|}
        	\hline 
        	&\includegraphics[width=.125\linewidth]{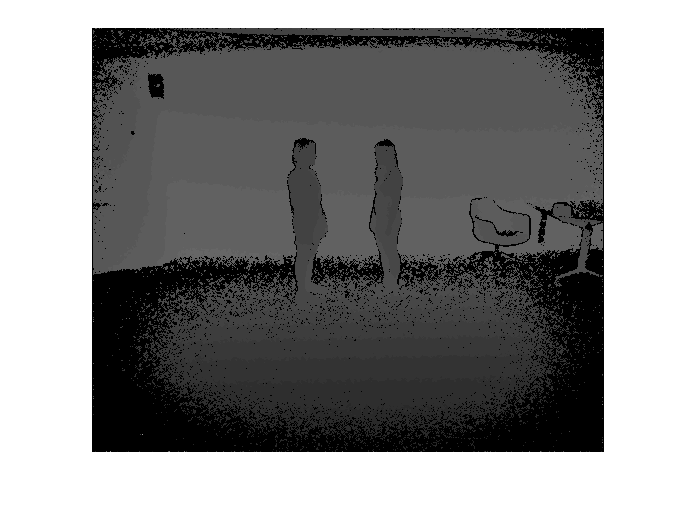}
			&\includegraphics[width=.125\linewidth]{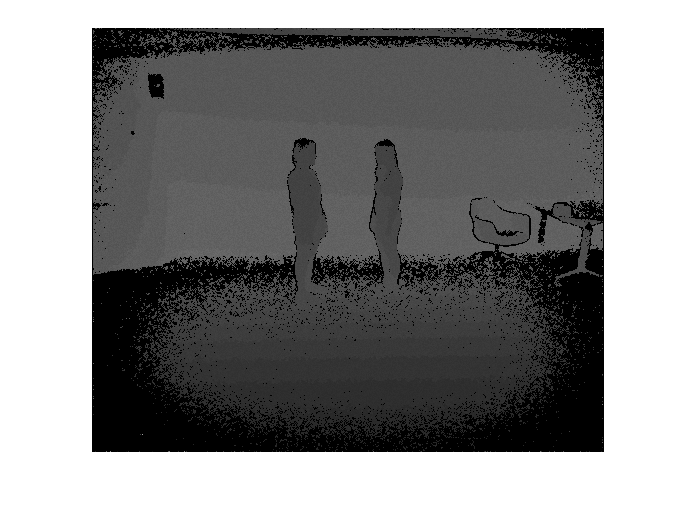}
            &\includegraphics[width=.125\linewidth]{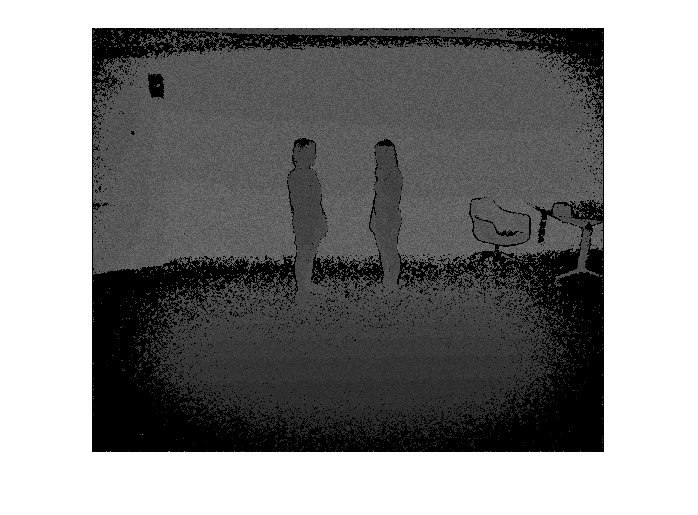}
            &\includegraphics[width=.125\linewidth]{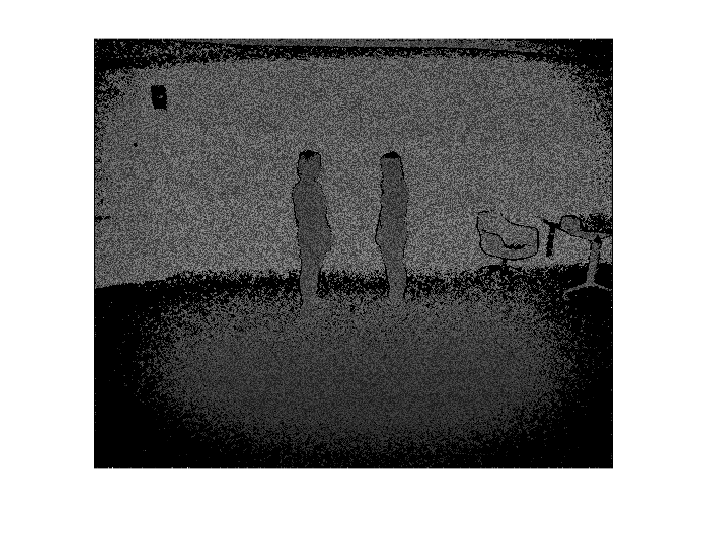}
            &\includegraphics[width=.125\linewidth]{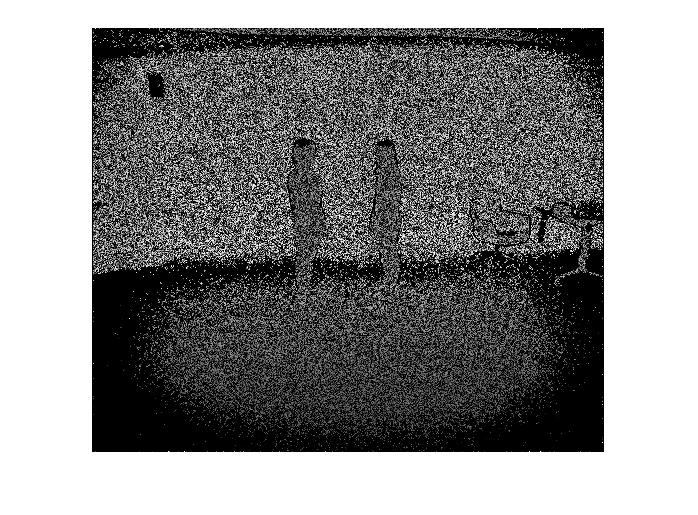}
            &\includegraphics[width=.125\linewidth]{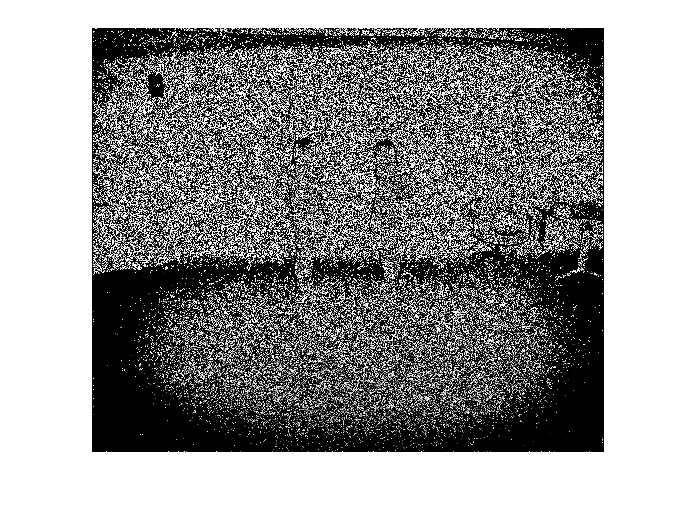}
            &\includegraphics[width=.125\linewidth]{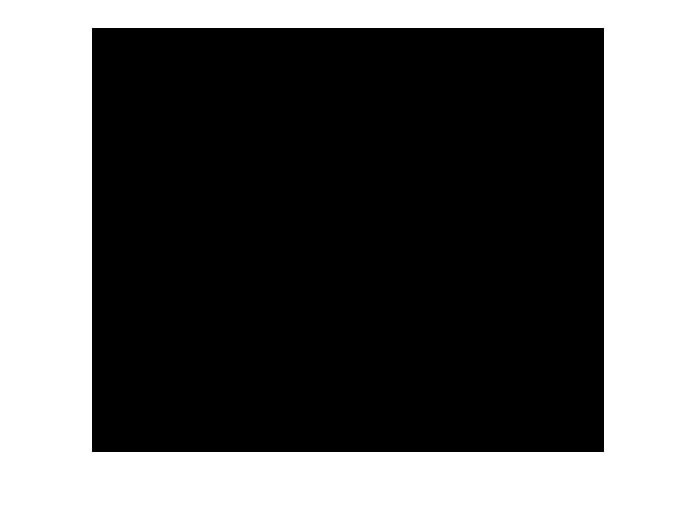}\\ 
            $\sigma^2$ & \textit{no noise} & $10^{-3}$ & $10^{-2}$& $10^{-1}$& $10^{0}$& $10^{1}$& \textit{void}\\ 
            Accuracy & 81.43\% & 81.34\% & 81.12\% & 76.85\% & 62.47\% & 51.43\% & 14.24\% \\ \hline            \end{tabular} 
   \caption{Accuracy of the model tested with clean RGB and noisy depth data. Accuracy of the proposed hallucination model, i.e. with \textit{no depth} at test time, is 77.21\%.}
       \vspace{-20pt}
   \label{tab:noise}
\end{table}
The depth sensor used in the NTU dataset (Kinect), is an IR emitter coupled with an IR camera, and has very complex noise characterization comprising at least 6 different sources \cite{kinect_noise}. It is beyond the scope of this work to investigate noise models affecting the depth channel, so, for our analysis we choose the most commonly adopted noise model, i.e., the multiplicative speckle noise.

Hence, we inject multiplicative Gaussian noise in the depth image $I$ in order to simulate speckle noise: $I=I * n, \, n \sim \mathcal{N}(1,\sigma) $. Table \ref{tab:noise} shows how performances of the network degrade when depth is corrupted with such Gaussian noise with increasing variance (cross-view protocol only). Results show that accuracy significantly decreases wrt the one guaranteed by our hallucination model (row \#15 in Table \ref{table:ablation}), even with low noise variance. This means, in conclusion, that training an hallucination network is an effective way not only to obviate to the problem of a missing modality, but also to deal with noise affecting the input data channel.

%%%%%%%%%%%%%%%%%%%%%%%%%%%%%%%%%%%%%%%%%%%%%%%%%%%%%%%%%%%%%%%%%
%%%%%%%%%%------------------------------------------%%%%%%%%%%%%%
%%%%%%%%%%%%%%%%%%%%%%%%%%%%%%%%%%%%%%%%%%%%%%%%%%%%%%%%%%%%%%%%%
\section{Conclusions and Future Work}\label{sec:concl}
In this paper, we address the task of video action recognition in the context of privileged information. We propose a new learning paradigm to teach an hallucination network to mimic the depth stream, yet receiving RGB as input. Our model outperforms many of the supervised methods recently evaluated on the NTU RGB+D dataset, as well as the hallucination model proposed in \cite{hoffman2016learning}. We conducted an extensive ablation study to verify how the several parts composing our learning paradigm contribute to the model performance. 

For future work, we would like to extend this approach to deal with additional modalities that may be available at training time, such as skeleton joints data or infrared sequences. 

\clearpage

\bibliographystyle{splncs}
\bibliography{egbib}

\begin{thebibliography}{10}

\bibitem{vapnik2009new}
Vapnik, V., Vashist, A.:
\newblock A new learning paradigm: Learning using privileged information.
\newblock Neural networks \textbf{22}(5) (2009)  544--557

\bibitem{hoffman2016learning}
Hoffman, J., Gupta, S., Darrell, T.:
\newblock Learning with side information through modality hallucination.
\newblock In: Proceedings of the IEEE Conference on Computer Vision and Pattern
  Recognition. (2016)  826--834

\bibitem{hinton2014distilling}
Hinton, G., Vinyals, O., Dean, J.:
\newblock Distilling the knowledge in a neural network.
\newblock Deep Learning and Representation Learning Workshop: NIPS 2014 (2014)

\bibitem{ba2014deep}
Ba, L.J., Caruana, R.:
\newblock Do deep nets really need to be deep?
\newblock Proceedings of Advances in Neural Information Processing Systems
  (NIPS) (2014)

\bibitem{lopez2015unifying}
Lopez-Paz, D., Bottou, L., Sch{\"o}lkopf, B., Vapnik, V.:
\newblock Unifying distillation and privileged information.
\newblock Proceedings of the International Conference on Learning
  Representations (ICLR) (2016)

\bibitem{simonyan2014two}
Simonyan, K., Zisserman, A.:
\newblock Two-stream convolutional networks for action recognition in videos.
\newblock In: Advances in neural information processing systems. (2014)
  568--576

\bibitem{carreira2017quo}
Carreira, J., Zisserman, A.:
\newblock Quo vadis, action recognition? a new model and the kinetics dataset.
\newblock In: Proceedings of the IEEE Conference on Computer Vision and Pattern
  Recognition. (2017)

\bibitem{feichtenhofer2017spatiotemporal}
Feichtenhofer, C., Pinz, A., Wildes, R.P.:
\newblock Spatiotemporal multiplier networks for video action recognition.
\newblock In: Proceedings of the IEEE Conference on Computer Vision and Pattern
  Recognition. (2017)  4768--4777

\bibitem{shahroudy2016ntu}
Shahroudy, A., Liu, J., Ng, T.T., Wang, G.:
\newblock Ntu rgb+ d: A large scale dataset for 3d human activity analysis.
\newblock In: Proceedings of the IEEE Conference on Computer Vision and Pattern
  Recognition. (2016)  1010--1019

\bibitem{luo2017graph}
Luo, Z., Jiang, L., Hsieh, J.T., Niebles, J.C., Fei-Fei, L.:
\newblock Graph distillation for action detection with privileged information.
\newblock arXiv preprint arXiv:1712.00108 (2017)

\bibitem{feichtenhofer2016convolutional}
Feichtenhofer, C., Pinz, A., Zisserman, A.:
\newblock Convolutional two-stream network fusion for video action recognition.
\newblock In: Proceedings of the IEEE Conference on Computer Vision and Pattern
  Recognition. (2016)  1933--1941

\bibitem{shahroudy2017deep}
Shahroudy, A., Ng, T.T., Gong, Y., Wang, G.:
\newblock Deep multimodal feature analysis for action recognition in rgb+ d
  videos.
\newblock IEEE Transactions on Pattern Analysis and Machine Intelligence (2017)

\bibitem{liu2017viewpoint}
Liu, J., Akhtar, N., Mian, A.:
\newblock Viewpoint invariant action recognition using rgb-d videos.
\newblock arXiv preprint arXiv:1709.05087 (2017)

\bibitem{he2016deep}
He, K., Zhang, X., Ren, S., Sun, J.:
\newblock Deep residual learning for image recognition.
\newblock In: Proceedings of the IEEE conference on computer vision and pattern
  recognition. (2016)  770--778

\bibitem{he2016identity}
He, K., Zhang, X., Ren, S., Sun, J.:
\newblock Identity mappings in deep residual networks.
\newblock In: European Conference on Computer Vision, Springer (2016)  630--645

\bibitem{eitel2015multimodal}
Eitel, A., Springenberg, J.T., Spinello, L., Riedmiller, M., Burgard, W.:
\newblock Multimodal deep learning for robust rgb-d object recognition.
\newblock In: Intelligent Robots and Systems (IROS), 2015 IEEE/RSJ
  International Conference on, IEEE (2015)  681--687

\bibitem{luo2017unsupervised}
Luo, Z., Peng, B., Huang, D.A., Alahi, A., Fei-Fei, L.:
\newblock Unsupervised learning of long-term motion dynamics for videos.
\newblock In: IEEE Conference on Computer Vision and Pattern Recognition
  (CVPR). Number EPFL-CONF-230240 (2017)

\bibitem{ohn2013joint}
Ohn-Bar, E., Trivedi, M.M.:
\newblock Joint angles similarities and hog2 for action recognition.
\newblock In: Computer vision and pattern recognition workshops (CVPRW), 2013
  IEEE conference on, IEEE (2013)  465--470

\bibitem{soo2017interpretable}
Soo~Kim, T., Reiter, A.:
\newblock Interpretable 3d human action analysis with temporal convolutional
  networks.
\newblock In: Proceedings of the IEEE Conference on Computer Vision and Pattern
  Recognition Workshops. (2017)  20--28

\bibitem{kinect_noise}
Mallick, T., Das, P.P., Majumdar, A.K.:
\newblock Characterizations of noise in kinect depth images: A review.
\newblock IEEE Sensors Journal \textbf{14}(6) (June 2014)  1731--1740

\end{thebibliography}


\begin{thebibliography}{1}

\bibitem{feichtenhofer2017spatiotemporal}
Feichtenhofer, C., Pinz, A., Wildes, R.P.:
\newblock Spatiotemporal multiplier networks for video action recognition.
\newblock In: Proceedings of the IEEE Conference on Computer Vision and Pattern
  Recognition. (2017)  4768--4777

\bibitem{shahroudy2016ntu}
Shahroudy, A., Liu, J., Ng, T.T., Wang, G.:
\newblock Ntu rgb+ d: A large scale dataset for 3d human activity analysis.
\newblock In: Proceedings of the IEEE Conference on Computer Vision and Pattern
  Recognition. (2016)  1010--1019

\end{thebibliography}
\end{document}

\pagestyle{headings}
\mainmatter
	
\title{Modality Distillation with Multiple Stream Networks for Action Recognition\\-\\Supplementary material}
	
\titlerunning{Modality Distillation with Multiple Stream Networks}
	
\author{Nuno C. Garcia \and
		Pietro Morerio \and
		Vittorio Murino}
	\institute{}
	
\maketitle

	In this document, we address in more details a few topics referred to in the main paper.
	Section \ref{sec:RGBD} explores how the model learns with inverted cross-stream connection, \textit{i.e.}, the depth stream network is receiving the signal from the RGB stream (and thus the RGB stream is the missing modality).
	We confirm the intuition that in this case RGB
	is not adding useful information to the depth stream network and, overall,  
	this leads to a lower performance of the model. 
	Next, we describe in section \ref{sec:detail} some implementation details that may be useful for reproducibility. 
	
	%%%%%%%%%%%%
	\section{Inverting the data modalities: RGB distillation}
	\label{sec:RGBD}
	Despite the proposed architecture is general and can be applied to any multimodal pair of data streams, our model is not symmetric under the swap of the depth and RGB modalities. In fact, the connection between streams is engineered such that the RGB stream is fed with a signal coming from the depth stream, and not vice versa.
	The intuition for such choice of direction is that the depth stream learns from cleaner, more representative data (foreground depth maps), and is able to inform the RGB stream where the action is taking place, practically working as an augmentation tool for those regions of the feature map.
	In fact, the depth stream alone performs better the the RGB alone
	
	In \cite{feichtenhofer2017spatiotemporal}, the authors tested different locations where to inject the optical flow signal (in- or out- the ResNet residual unit). Bi-directional connections were also investigated (\textit{i.e.} both streams were injected one into the other). It was concluded that injecting signal into the optical flow stream decreases the model performance, and suggest that the reason can be ascribed to the RGB stream becoming dominant during training. We hypothesize that the same reasoning can be applied to the depth stream, which in our model takes the place of optical flow. However, in \cite{feichtenhofer2017spatiotemporal}, the authors did not try to invert the connection, \textit{i.e.} to inject signal from RGB to optical flow. 
	We report the results of such experiment in Table \ref{table:ablation}.
	\\
	
	\textbf{Discussion.} Line \#8a reports the accuracy obtained by the teacher network at the end of step 2: not only such accuracy is lower than the one of our original teacher network (line \#8), but also is only marginally higher than the one obtained by the final model (line \#15), which only uses RGB  at test time. Line \# 8a represents thus a very poor upper bound (as compared to line \# 8). 
	This translates in a worse hallucination network (lines \#13a) and worse distilled model (\#15a).
	
\begin{table}
	\centering
	\begin{center}
		\begin{tabular}{|ll|l|c|c|c|}
			\hline
			\# & Method & Test Modality & Loss &  Cross-Subject & Cross-View \\
			\hline
			\hline
			1 &Ours - step 1, depth stream & Depth &x-entr& 70.44\% & 75.16\% \\
			2 &Ours - step 1, RGB stream   & RGB   &x-entr& 66.52\% & 71.39\% \\
			\hline \hline
			\multicolumn{6}{|c|}{Depth $\rightarrow$ RGB (\textit{compare to Table 2 of the paper})} \\
			\hline
			6&Ours - step 2, depth stream & Depth &x-entr& 71.09\% & 77.30\% \\
			7&Ours - step 2, RGB stream & RGB &x-entr& 66.68\% & 56.26\% \\
			8&Ours - step 2 & RGB \& Depth &x-entr& \textbf{79.73}\% & \textbf{81.43}\% \\
			13&Ours - step 3 & RGB (\textit{hall}) & eq. (7)& 71.93\% & 74.10\% \\
			15&Ours - step 4 & RGB  &x-entr& \textbf{73.42}\% & \textbf{77.21}\% \\              
			\hline \hline
			\multicolumn{6}{|c|}{\textbf{Inverted - RGB $\rightarrow$ Depth}} \\
			\hline
			6a&Ours - step 2, depth stream & Depth &x-entr& 66.6\% & 73.68\% \\
			7a&Ours - step 2, RGB stream & RGB &x-entr& 63.98\% & 61.18\% \\
			8a&Ours - step 2 & RGB \& Depth &x-entr& \textit{74.45}\% & \textit{78.55}\% \\
			13a&Ours - step 3 & RGB (\textit{hall}) & eq. (7)& 68.47\% & 72.77\% \\
			15a&Ours - step 4 & RGB  &x-entr& \textit{66.86}\% & \textit{73.34\%} \\
			\hline          
		\end{tabular}
	\end{center}
	\caption{Inverting the cross-stream connection study. The last section of the table refers to results where the direction of the cross-stream connection has been inverted. The other results are also reported in the paper, as they refer to the model proposed.}
		\label{table:ablation}
	\end{table}

	\section{Implementation details}
	\label{sec:detail}
	\subsubsection{Pre-processing \& alignment.}
	The multiplicative cross-stream connections present in our model require both RGB and depth frames to be spatially aligned, since they are element-wise operations over the feature maps. Such alignment comes for free when using RGB and optical flow - which is computed directly from the appearance frames. However, this is not normally the case when using depth and RGB frames that are acquired with different sensors, and have different dimensions and aspect ratios as in the NTU RGB+D dataset, or other Kinect-acquired data. Fortunately, this dataset provides the joints' spatial coordinates in every RGB and depth frames, $rgb_{x,y}$ and $depth_{x,y}$ respectively, which we use to align both modalities.
	For every frame of a given video, we first compute the ratio $ratio_{x}^{A,B} = (rgb_{x}^{A} - rgb_{x}^{B}) / (depth_{x}^{A} - depth_{x}^{B}) \forall A,B \in S$, using all depth and RGB $x$ coordinates from the frame's well-tracked joints set $S$, and similarly for the $y$ dimension.
	The video aspect ratio is then calculated as the mean between the median aspect ratio for $x$ and the median aspect ratio for $y$ dimensions.
	The RGB frames of a given video are scaled according to this ratio. Finally, both RGB and depth frames are overlaid by aligning both skeletons, and the intersection is cropped on both modalities. The cropped sections are then rescaled according to the network's input dimension, in this case 224x224.
	Similarly to what was done in \cite{feichtenhofer2017spatiotemporal}, we sample 5 frames evenly spaced in time for each video, both for training and testing. For training, we also flip horizontally the video frames with probability $P=0.5$.
	
	%%%%%%%%%%%%
	\subsubsection{Hyperparameters and validation set.}
	After validation, we have selected the following set of hyperparameters: $\alpha$ = 0.5, $\lambda$ = 0.5, $T$ = 10.
	The validation set is not defined in the original paper where the dataset is presented \cite{shahroudy2016ntu}. For the sake of experiments reproducibility, we explain here how we defined the validation set. For the cross-subject protocol, we choose the subject \#1 (from the training set), which corresponds to around 5\% of the training set. For the cross-view protocol, we do the following: 1) create a dictionary of sorted videos for each key=action (from the training set); 2) set numpy random seed equal to 0; 3) sample 31 videos using \texttt{numpy.random.choice} for each action, which in the end will correspond to around 5\% of the training set.

		\bibliographystyle{splncs}
		\bibliography{egbib}